\documentclass[review]{elsarticle}

\usepackage{lineno,hyperref}
\usepackage{graphics}
\graphicspath{}
\usepackage{amsmath}
\usepackage{verbatim}
\usepackage{enumerate}
\usepackage{color}

\usepackage{algorithm}  
\usepackage{algpseudocode}  
\usepackage{amsmath}  
\usepackage{caption} 
\usepackage{array}
\usepackage{txfonts} 
\usepackage{ragged2e}
\usepackage{url}
\usepackage{amssymb}
\usepackage{array} 

\usepackage{booktabs} 

\usepackage{longtable}

\usepackage{bm}

\usepackage{setspace} 

\usepackage{natbib} 

\begin{document}
\captionsetup[figure]{labelfont={bf},name={Fig.},labelsep=period}
\captionsetup[table]{labelfont={bf},name={Table.},labelsep=period}
\begin{frontmatter}

\title{AE-Net: Autonomous Evolution Image Fusion Method Inspired by Human Cognitive Mechanism}
\author{Aiqing Fang, Xinbo Zhao*, Jiaqi Yang, Shihao Cao,Yanning Zhang}%
\address{
	\justifying\let\raggedright\justifying
National Engineering Laboratory for Integrated Aero-Space-Ground-Ocean Big Data Application Technology, School of Computer Science, Northwestern Polytechnical University, Xi’an 710072, China

}



\cortext[mycorrespondingauthor]{Corresponding author}
\ead{xbozhao@nwpu.edu.cn (X. Zhao); aiqingf@mail.nwpu.edu.cn (A. Fang);
	jqyang@nwpu.edu.cn;
	 ynzhang@nwpu.edu.cn}

\begin{abstract}
In order to solve the \textit{robustness and generality problems} of the image fusion task, inspired by the human brain cognitive mechanism, we propose a \textit{robust and general image fusion method with autonomous evolution ability, and is therefore denoted with AE-Net}. Through the collaborative optimization of multiple image fusion methods to simulate the cognitive process of human brain, unsupervised learning image fusion task can be transformed into semi-supervised image fusion task or supervised image fusion task, thus promoting the evolutionary ability of network model weight. \textbf{Firstly}, the relationship between human brain cognitive mechanism and image fusion task is analyzed and a physical model is established to simulate human brain cognitive mechanism. \textbf{Secondly}, we analyze existing image fusion methods and image fusion loss functions, select the image fusion method with complementary features to construct the algorithm module, establish the multi-loss joint evaluation function to obtain the optimal solution of algorithm module. The optimal solution of each image is used to guide the weight training of network model. Our image fusion method can effectively unify the cross-modal image fusion task and the same modal image fusion task, and effectively overcome the difference of data distribution between different datasets. \textbf{Finally}, extensive numerical results verify the effectiveness and superiority of our method on a variety of image fusion datasets, including multi-focus dataset, infrared and visible dataset, medical image dataset and multi-exposure dataset. \textit{Comprehensive experiments demonstrate the superiority of our image fusion method in robustness and generality.} In addition, experimental results also demonstate the effectiveness of human brain cognitive mechanism to improve the robustness and generality of image fusion. 
\end{abstract}

\end{frontmatter}

{A}{s} a basic task in the field of computer vision, image fusion plays an important role in visual navigation, object detection and image caption. In the task of image fusion, the robustness and generality of image fusion in complex environment has always been a bottleneck problem that puzzles and restricts the application of technology. However, \textit{human brain has strong robustness and generality for various computer vision tasks, which are closely related to the cognitive processing mechanism of human brain}. \textit{According to the research of cognitive psychology \cite{MillerCPortex} and biological neuroscience \cite{GuangYang2009Smds}, human brain has the ability of working memory and continuous learning.} In contrast, human beings constantly extract knowledge, modify experience and store it in the form of working memory in daily life through their own experience. The extraction and storage of this knowledge will greatly promote the working ability of human beings. When dealing with new tasks, human beings will establish the difference model between prior knowledge of existing tasks and new tasks according to the existing working memory, and modify prior knowledge according to the differences to solve new tasks. At the same time, based on the understanding of the task, human beings will decompose the task continuously, and seek the optimal solution of the decomposed subtask, so as to promote the global optimal solution of the task. It is the human brain's task-oriented cognitive processing mechanism that makes the human brain has strong robustness and versatility in processing many visual tasks, without the need to remodel each new task. \textit{Therefore, we believe that the cognitive learning process of the human brain has positive significance to improve the robustness and universality of image fusion task, as verified in Sect.4.} 

In the task of image fusion, \textit{many image fusion methods are proposed based on the characteristics of human visual system}. For example, deep convolution neural network image fusion method based on biological neuroscience \cite{9031751, PGMI, xu2020aaai, DengXin2020DCNN, Liu2017MultiCNN, MaFusionGAN, Yin2018MedicalNSSPAPCNN, ZHANG202099, fang2019crossmodale, FangAiqing2020NaSF}, visual saliency image fusion method based on the human visual attention mechanism \cite{Bavirisetti2016Two, Zhang2015A, Cui2015Detail, ZhaoT.2019Pfan, fang2019crossmodale, Lahoud2019FastZERO}, multi-task collaborative optimization image fusion method based on the auxiliary characteristics of human brain \cite{fang2019crossmodal}, and image fusion method based on the selective characteristic and the nonlinear fusion characteristic of human visual system \cite{FangAiqing2020NaSF} et al. These methods have explored the different aspects of human brain and human visual system in the field of image fusion, and achieved significant research results. To improve the robustness and generality of image fusion, some \textit{general network frameworks} \cite{PGMI, xu2020aaai, ZHANG202099} are proposed. \textit{These methods include three parts.} The above methods improve the robustness and generality of image fusion to a certain extent. Nonetheless, \textit{there is a big gap between the robustness and generality of human brain visual system.} \textbf{Firstly}, these methods donot consider the problem of memory forgetting due to the difference of data distribution. \textbf{Secondly}, the complementarity of different methods is not considered. \textbf{Finally}, \textit{existing image fusion methods have paid few research attention on human brain cognitive processing mechanism.} By contrast, more researches have been conducted from the technology perspective.

\begin{figure}[ht]
	\centering
	\includegraphics[width=0.48\textwidth]{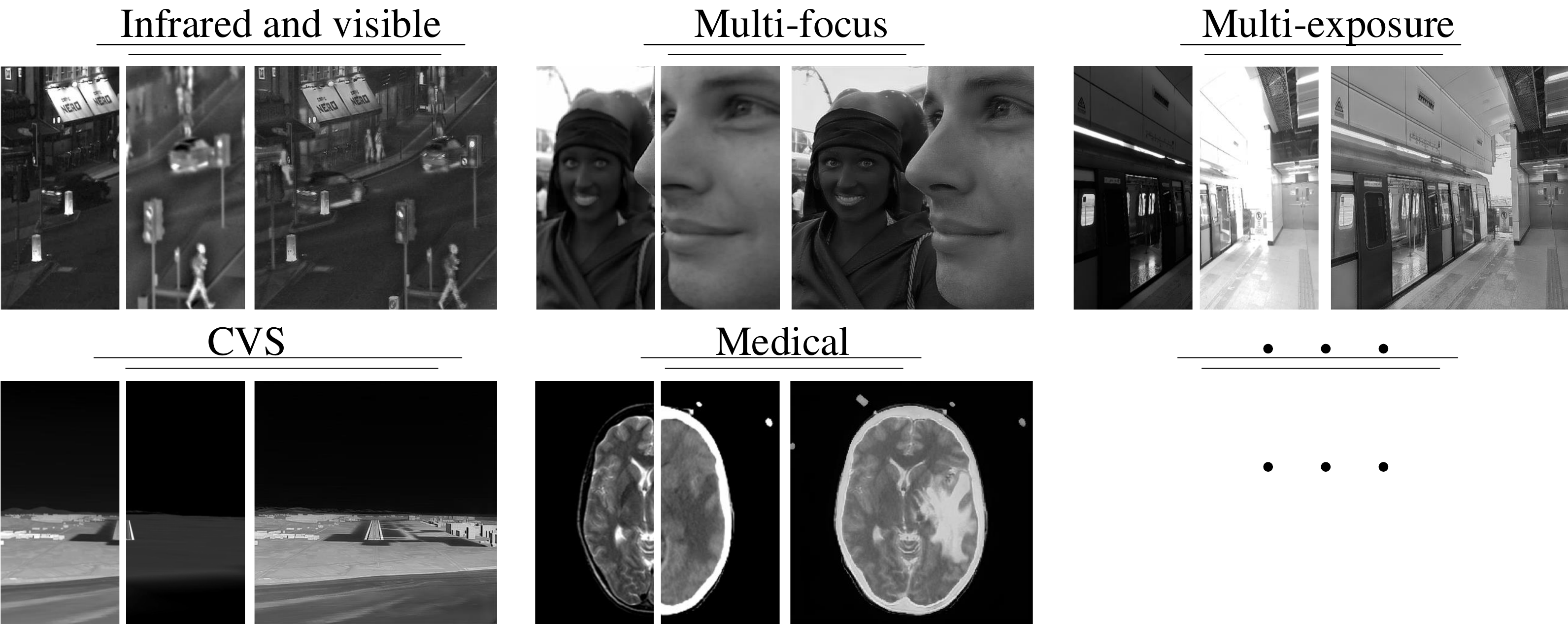}
	\caption{Examples of different image fusion tasks. Including infrared and visible image fusion, multi-focus image fusion, multi-exposure image fusion, CVS image fusion, medical image fusion, et al.\label{summary}}
\end{figure}

In order to overcome above problems, inspired by the human brain cognitive mechanism, we propose a robust and general image fusion method with autonomous evolution ability. \textit{As far as we know, this is the first image fusion method with autonomous learning and evolution function in the field of image fusion.} Our image fusion method mainly includes \textit{three modules}, including multi-method collaborative module, multi-index evaluation module and iterative learning optimization module. The initial fusion result of image fusion task is obtained by multi-method cooperation; the optimal fusion data of existing methods is evaluated by multi-index evaluation function of image quality; the weight optimization learning is carried out by iterative optimization module on the basis of obtaining the optimal fusion solution of each step. \textbf{On the one hand}, our image fusion method can solve the problem of modeling the loss function of cross-modal image fusion task. \textbf{On the other hand}, it can simulate the cognitive processing mechanism of human brain, so that the network can integrate a variety of image fusion methods. Our method is based on modular design, which makes the algorithm has strong scalability. We can replace the algorithm module to adapt to different data fusion tasks and improve the fusion effect of the network. \textit{Our image fusion method is not a simple superposition of existing image fusion methods, whereas a combination of human task-oriented processing mechanism and image fusion task to simulate the cognitive processing mechanism of human brain, improve the robustness and generality of image fusion, and make the network architecture have the ability of continuous learning and evolution.} The main contributions of our work include the following three points:
\begin{itemize}
	\item \textbf{Firstly}, we explored the close relationship between the cognitive processing mechanism of human brain and image fusion task, and established a physical model to simulate the cognitive mechanism of human brain.
	
	\item \textbf{Secondly}, based on the cognitive mechanism of human brain, we propose a general image fusion method and establish a general image fusion network framework with continuous learning ability. Our image fusion method has the ability of autonomous evolutionary learning and can be compatible with all existing image fusion algorithms.
	
	\item \textbf{Finally}, hybrid dataset benchmark. \textit{Due to the lack of effective data benchmark in the evaluation of image generality and robustness, the evaluation of algorithm quality becomes a blind spot}. Image scene includes highlight, dark light, blur, noise and so on. We have collected and produced a dataset from existing datasets, which involves multi-exposure images, multi-focus images, infrared and visible images, remote sensing images, medical images. \textit{We will release our hybrid benchmark dataset after paper is accepted}. In this benchmark, we will provide fused image and evaluation codes of all test images of 29 image fusion methods on my \href{https://github.com/AiqingFang/Image-Fusion-Summary}{GitHub} homepage. It provides a baseline for the comparison and reference of robustness and generality of existing image fusion methods.

\end{itemize}
The remainder of this paper is structured as follows. Sect. 2 reviews relevant theory knowledge. Sect. 3 presents a robust and general image fusion method. Sect. 4 introduces the experimental datasets, evaluation metrics, and implementation details. Sect. 5 presents a discussion and explanation. Sect. 6 gets a conclusion.

\section{Related work}
Our research content includes two aspects: human brain cognitive mechanism and image fusion.
\subsection{Human Brain Cognitive Mechanism}
In recent years, the deep learning method inspired by biological neuroscience has made remarkable achievements in many fields of computer vision. However, compared with human brain, there is a great gap in power consumption, generality, robustness and continuous learning ability. Therefore, researchers have carried out in-depth research on human brain cognitive mechanism. In the process of biological experiments, Yang et al. \cite{GuangYang2009Smds} proposed that the postsynaptic spines learning and new sensory experiences of mouse cortex 7 and 8 will lead to the formation and elimination of spine through a protected process. \textit{This study suggests that continuous learning in the neocortex depends on task-specific synaptic consolidation, in which knowledge is encoded persistently by making some synapses less plastic, so it can be stable for a long time}. Miller et al. \cite{MillerCPortex, KirkpatrickJames2017Ocfi} pointed out that \textit{prefrontal cortex is responsible for the integration of human brain's perception, cognitive reasoning, continuous learning, working memory storage and other information}. It is the ability of working memory and continuous learning of human brain that enables human beings to rapidly and continuously increase their knowledge and ability without a lot of data driven. However, this process of autonomous evolutionary learning is deficient in the existing deep learning methods \cite{Cvl20}. To solve this problem, this ability of human brain has been simulated in the field of computer vision \cite{Zohary1992Population}, and some achievements have been made. For example, progressive neural network \cite{RusuAndrei2016PNN}, path neural network \cite{BanarseDylan2017PECG}, ewc \cite{KirkpatrickJames2017Ocfi}, etc. However, there is still a big gap between these methods and human brain, which needs further research. In the field of image fusion, \textit{no more research results related to human brain cognitive mechanism have been found}. 
\subsection{Image Fusion}
We classify image fusion methods into traditional ones and deep-learning-based ones.

\textit{1) Image fusion method based on traditional method}. 
To solve the problem of moving ghost in multi-exposure fusion task, Qin et al. \cite{XiamengQin2015RMFU} proposed patch based match and fusion method. This is the first patch based exposure fusion method to preserve the moving objects of dynamic scenes that does not need the registration process of different exposure images. Shen et al. \cite{JianbingShen2014EFUB} first evaluated the multi-exposure image quality measurement and human vision system, and constructed a hybrid multi-exposure fusion method. Durga et al. \cite{Bavirisetti2016Two} combines two scale decomposition with visual saliency for image fusion. Wei et al. \cite{GanWei2015Iavi} evaluated the effectiveness of multi-scale edge preserving decomposition and guided image filter. Zhao et al. \cite{Zhao2017Multisensor} proposed a new fusion framework that integrates image fusion based on spectral total variation (TV) method and image enhancement.  Li et al.\cite{Li2018InfraredLTLRR}proposed a multi-level image decomposition method based on latent lowrank representation in image fusion task. Fayez et al. \cite{Lahoud2019FastZERO} proposed to combine the tow scale, visual saliency and deep features to evaluate the impact on image fusion quality. The above image fusion methods have achieved some results, whereas the characteristics of human visual system are not considered.
Although Fang et al. \cite{fang2019crossmodale} first evaluated the influence of human visual characteristics on the image fusion task. Nonetheless, above image fusion methods are more from the image fusion technology itself, lack of related research on human brain cognitive processing mechanism.

\textit{2) Image fusion method based on deep learning}. 
Liu et al. \cite{Yu2017medicalCNN} first proposed a medical image fusion method based on deep learning and infrared and visible image fusion, and extracted the deep layer by deep convolution neural network. Prabhakark et al \cite{PrabhakarK.Ram2017DADU}. proposed an unsupervised deep learning method for multi-exposure image fusion for the first time. Ma et al. \cite{MaFusionGAN} introduced the countermeasure generation network into the image fusion task for the first time, and optimized the image fusion weight through the confrontation loss, whereas this method has the problem of fusion ambiguity. To solve this problem, Ma et al. \cite{MA202085} proposed a infrared and visible image fusion via detail preserving transverse learning. Zhang et al. \cite{ZHANG202099} first proposed a general framework for image fusion, which effectively integrated the maximum fusion, sum fusion and weighted average fusion criteria. Xu et al. \cite{xu2020aaai} proposed a unified dense connected network for image fusion. This method effectively combines unsupervised learning with dense link network. Compared with the dense network fusion method proposed by Liu \cite{Yu2017medicalCNN}, this method can update the fusion weight. Ma et al. \cite{PGMI} proposed a fast and general image fusion network, which can extract features through the loss of image gradient and contrast, and effectively avoid the problem of information loss through feature reuse. Recently, Deng et al. \cite{DengXin2020DCNN} proposed a novel deep convolutional neural network to solve the general multi-modal image restoration (MIR) and multi-modal image fusion (MIF) problems. However, there are two problems in the above methods. \textbf{Firstly}, like traditional image fusion algorithms, the above methods are more improved than simultaneous interpreting algorithms, ignoring the close relationship between image fusion and human brain cognitive process. \textbf{Secondly}, it is difficult to model the task loss function of cross-modal image fusion, which is lack of further research. To solve this problem, Fang et al. \cite{fang2019crossmodale} has carried on the related research based on the human visual system's multi-task assisted learning chacteristic, and has proved that this feature can improve the robustness of image fusion. \textit{Although the above methods have achieved certain results, there is a lack of further research on the cognitive mechanism of human brain.}

In conclusion, inspired by the human brain task-oriented processing mechanism, an image fusion method with continuous learning ability is proposed, and a robust and general image fusion network architecture is designed. \textit{Our image fusion method can improve the robustness and generality of image fusion by introducing multi-method collaborative evaluation module to simulate human task-oriented processing mechanism.}

\section{Method}
Our image fusion method includes three steps. \textbf{Firstly}, the cognitive mechanism of human brain and the task of image fusion are analyzed and the physical model is established. \textbf{Secondly}, image fusion methods with complementary characteristics and common image fusion loss functions are analyzed. \textbf{Finally}, we build a general continuous learning image fusion network.
\begin{figure}[ht]
	\setlength{\belowcaptionskip}{-0.5cm}
	\centering
	\includegraphics[width=0.49\textwidth]{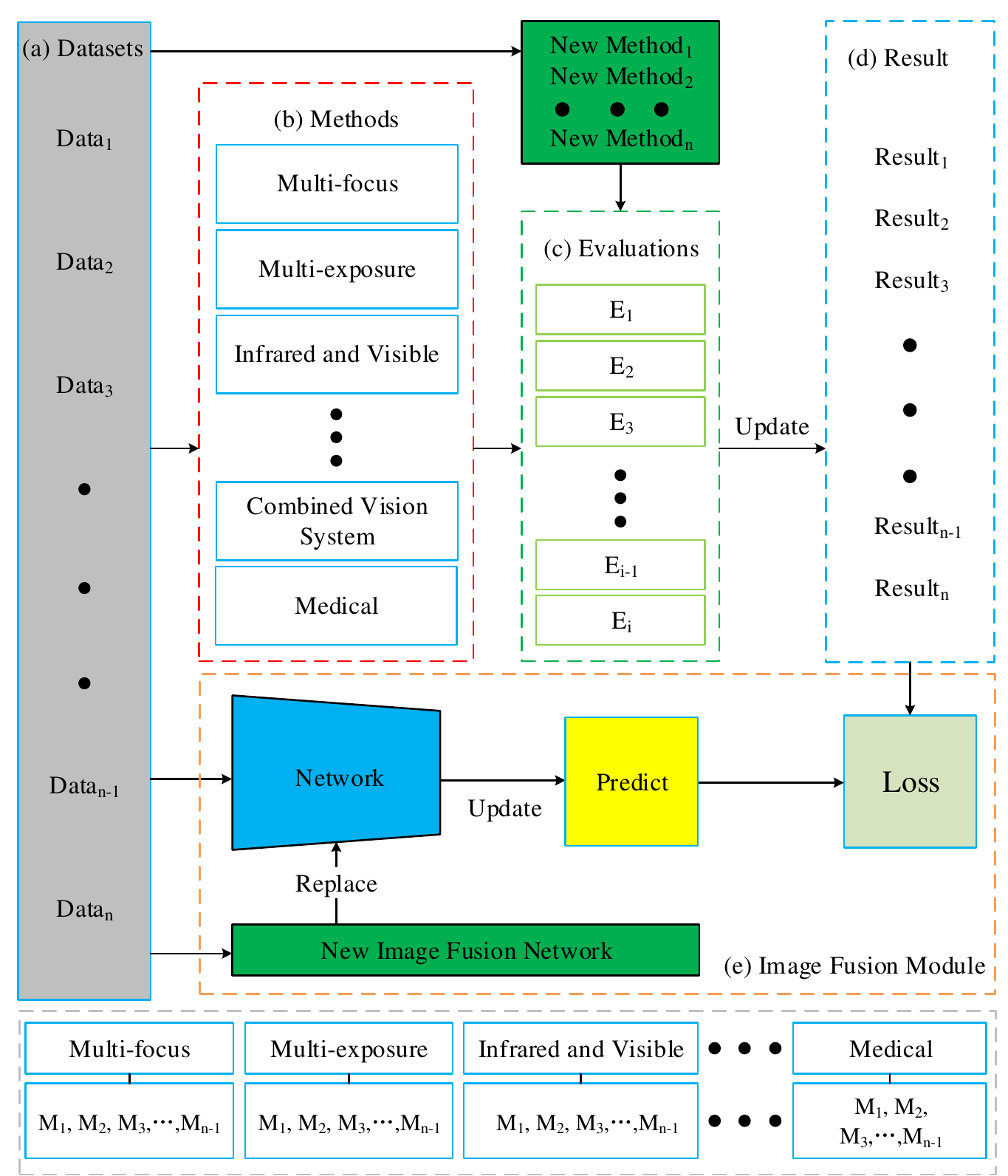}
	
	\caption{Network Architecture of the proposed AE-Net, where $E_1$ represents supervised learning image quality assessment function; $E_2$ represents unsupervised learning image quality assessment function; $Loss$ represents the loss of image fusion task. (a) represents dataset module. (b) represents different image fusion methods module. (c) Image quality evaluation module. (d) Initial optimal solution module. (e) Image fusion module.
	}
	\label{general}
\end{figure}
\subsection{Human Brain Cognitive Mechanism and Image Fusion}
In this section, we first represent the motivation of introducing human cognitive mechanism into image fusion task. Then, we introduce the physical modeling method of human cognitive mechanism into image fusion task.

\subsubsection{Human Brain Cognitive Mechanism Motivation}
Existing image fusion tasks, including multi-exposure image fusion task, infrared and visible image fusion task, multi-focus image fusion task, medical image fusion task and remote sensing image fusion task have the differences of imaging attributes and data distribution. These differences lead to the lack of robustness and generality of existing image fusion methods. Although some general network frameworks have been proposed, they are robust only in a limited number of image fusion tasks. However, \textit{the human brain is not affected by this difference and has robustness in multiple image fusion tasks}. The main reason is that the \textit{existing network methods lack the ability of autonomous evolutionary learning of human brain, and do not have the processing mechanism for different tasks. If there is an image fusion method that can update its network weight with new tasks or new methods as the human brain, without losing the existing prior knowledge, then the method has the function of autonomous learning and evolution of human brain}. This method will be the same as the human brain, the robustness and versatility of image fusion will no longer be affected by imaging attributes and data distribution differences. The details of AE-Net are described in the next subsection.

\subsubsection{Cognitive Mechanism Modeling of Human Brain}
In order to make the image fusion method robust on multiple image fusion datasets, we model the autonomous evolution ability of human brain. Image fusion method module $F_i$ is defined as:

\begin{equation}
\begin{array}{ll}
Fi=\{ME(x, y)_1, {V}({x}, {y})_2, {MF}({x}, {y})_3, \ldots, {M}({x}, {y})_{i}\}, \\ {i}=\{1,2, \ldots, {n}\},
\end{array}
\end{equation}
where $ME(x, y)$ indecates multi-exposure image fusion method; $V(x, y)$ represents infrared and visible image fusion method; $MF(x, y)$ represents multi-focus image fusion method; $M(x, y)$ represents medical image fusion method. Each kind of image fusion method in $F_i$ is defined as follows:

\begin{equation}\begin{array}{l}
\begin{aligned}
{ME}({x}, {y})&=\{{ME}_1, {ME}_2, \ldots, {ME_n}\} \\
{V}({x}, {y})&=\{{V}_1, {V}_2, \ldots, {V_n}\} \\
{MF}({x}, {y})&=\{{MF}_1, {MF}_2, \ldots, {MF_n}\} \\
{M}({x}, {y})&=\{{M}_1, {M}_2, \ldots, {M_n}\} \\
\end{aligned}

\end{array}\end{equation}

For $F_i$, the image quality evaluation function e of image fusion task should be established to evaluate the image quality of different datasets. The image quality evaluation function $E$ is defined as:
\begin{equation}{E}=\{{E}_1, {E}_2, {E}_3, \ldots, {E}_{i-1}, {E_i}\}, {i}=\{1,2, \ldots, {n}\},\end{equation}
where $E_i$ \ref{Eq1} \ref{Eq2} represents the evaluation function corresponding to the i-th image fusion method. The evaluation function is used to obtain the relative optimal solution $O$ of each image fusion effect in the dataset, $O$ is defined as:
\begin{equation}
{O}=\{Result_1, Result_2 , \ldots, Result_n \}, {i}={1,2, \ldots, n},\end{equation}
where $O$ is used to guide and optimize each iteration in the neural network model; $Result_i=Ei(p_i)$. By selecting the local optimal label data as the optimal solution of the deep neural network, the objective function $L(x, y)$ can be transformed from unsupervised learning method to semi-supervised or supervised learning method $L_p (x, y)$. For unsupervised learning, $L(x, y)$ is defined as:

\begin{equation}
{L}({x}, {y})=Argmin({E}({x}, {p})+{E}({y}, {p})) / 2.0,
\end{equation}
where $p$ represents image fusion result of neural network; $Lp(x,y)$ is defined as:

\begin{equation}
\begin{array}{ll}
\begin{aligned}
Lp(x,y)&=Argmin{E}({p}, {O})+C[L(x,y),0],  \\ 
\end{aligned}
\end{array}
\end{equation}
where $C[L(x,y),0]$ indicates that the selection function $C[]$ selects semi-supervised learning or supervised learning. By minimizing the objective function $L_P (x, p)$, the weights of network parameters are optimized. When the network architecture needs to add a new algorithm, we only need to find the optimal solution $O$ and update the data parameters of the optimal solution $O$ with the results of the new image fusion method $p_{new}$. The updated optimal solution is expressed as $O=Ei(p_{new},O)$. The updated optimal solution $O$ can be used to optimize the processing of the new image fusion task.

\subsection{Multiple Image Fusion Methods and Image Quality Evaluation}
\begin{figure}[ht]
	\setlength{\belowcaptionskip}{-0.5cm}
	\centering
	\includegraphics[width=0.45\textwidth]{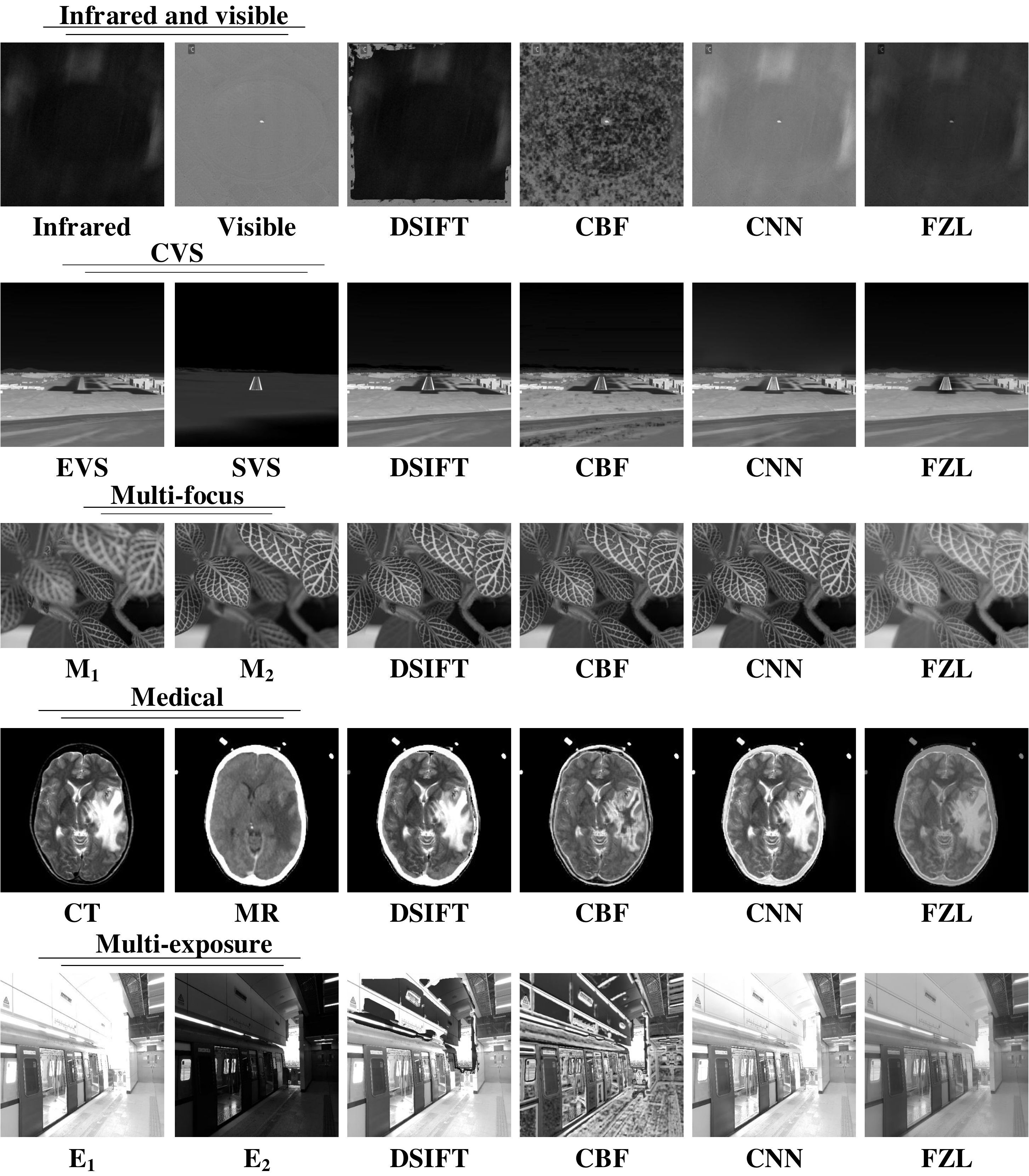}
	\caption{Comparison of image fusion methods for different tasks.}
	\label{generalm}
\end{figure}

In the existing image fusion method, each method has its own advantages, each method also has its defects. At present, no algorithm has been found to be robust in complex scenes such as multi-exposure image fusion task, infrared and visible image fusion task, multi-focus image fusion task, medical image fusion task, combined vision system image fusion task et al. In order to show the problems of existing methods, we make quantitative comparative analysis on different datasets of existing image fusion methods. As shown in Figure \ref{generalm}, we can see that traditional image fusion methods and latest image fusion methods based on deep learning cannot achieve robustness on multiple datasets at the same time. Inspired by the characteristics of human brain, an image fusion method with autonomous evolutionary learning ability is proposed. Our method needs to establish different image quality assessment methods for different datasets. In this paper, we mainly divide image quality assessment methods into labeled image quality assessment model and unlabeled image quality assessment model. For supervised image quality assessment, SSIM \cite{1284395} and PSNR \cite{SijbersJ1996Qaio} are commonly used in image quality assessment. Therefore, we use $S(p,r)$ and $P(p,r)$ as image quality assessment indicators of supervised learning. Image quality assessment function $E_1$ is defined as:

\begin{equation}\label{Eq1}{E}_1=\beta \times {S}({p}, {r})+\beta_1 \times {P}({p}, {r}),\end{equation}
where $E_1$ is the object evaluation function of supervised learning image fusion; $\beta$ and $\beta_1$ indicate the weight of $S(p,r)$ and $P(p,r)$.
\begin{figure}[ht]
	\setlength{\belowcaptionskip}{-0.5cm}
	\centering
	\includegraphics[width=0.48\textwidth]{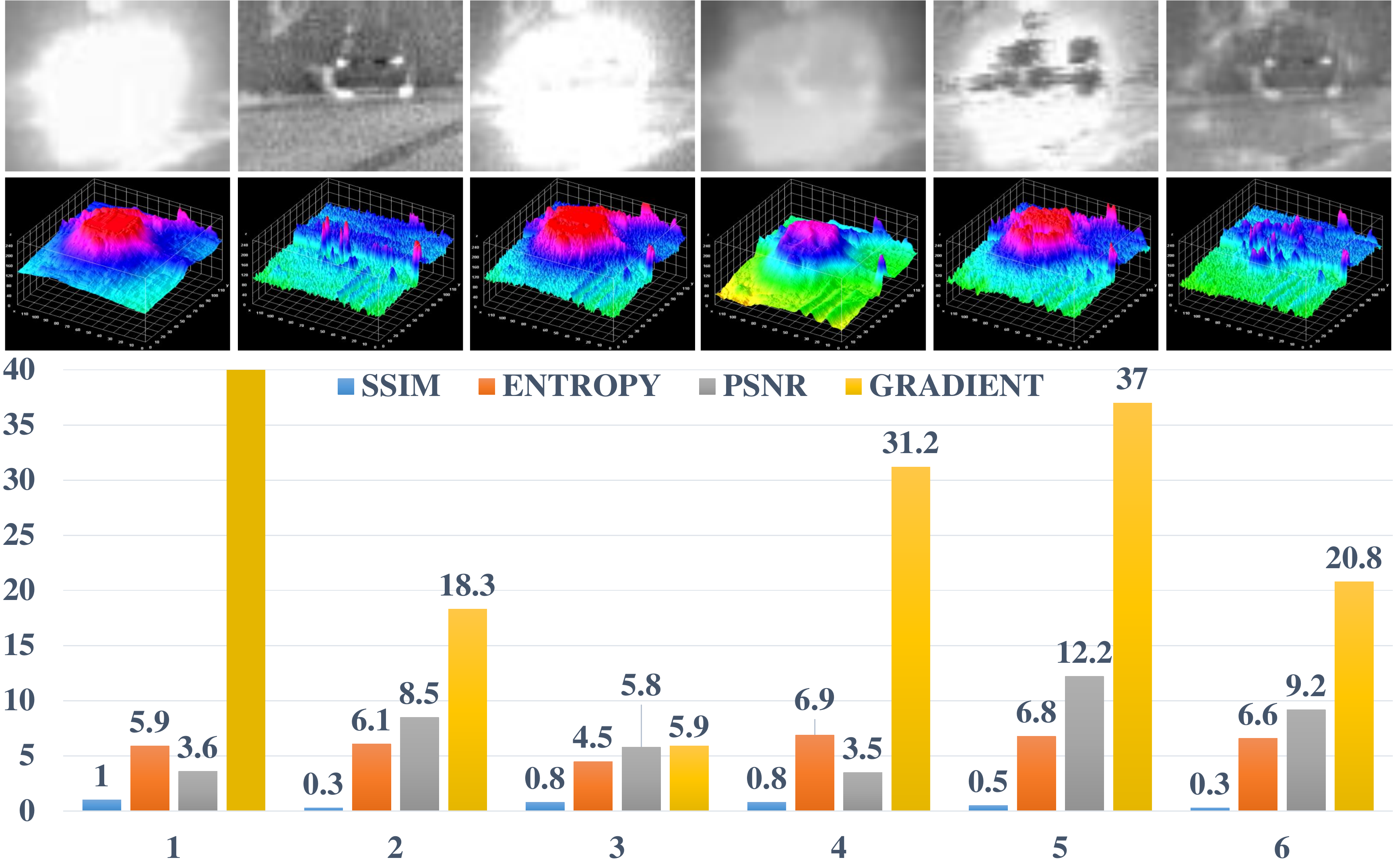}
	
	\caption{Comparison of image objective index and subjective quality. From top to bottom are image, light intensity and objective index. The objective indexes from left to right were SSIM \cite{1284395}, ENTROPY \cite{1576816}, PSNR \cite{SijbersJ1996Qaio} and Average gradient \cite{Cui2015Detail}.}
	\label{Fig4}
\end{figure}
However, it is very difficult to evaluate the image quality completely for the cross-modal image quality assessment task without ground truth labels. A single objective quality evaluation index has been unable to effectively characterize the image quality \cite{Valois1996Visual}. As shown in Figure \ref{Fig4}, we can see the defects of single index image quality assessment method. Therefore, we propose a multi-objective cross-modal image quality assessment method. Our cross-modal image quality assessment function $E_2$ combines natural image quality evaluator index $N(p)$, structural similarity objective function $S(p, r)$, information entropy index $G(p)$, energy gradient $B(p)$, visual fidelity $V(p, r)$, peak signal-to-noise ratio index $P(p, r)$ to evaluate image quality. Our cross-modal image quality assessment function $E_2$ is defined as:
\begin{equation}
\label{Eq2}
\begin{array}{ll}
{E}_ 2={\alpha \times N(p)}+{\alpha_1 \times G(p)}+ {\alpha_2 \times V(p,r)}+  
{\alpha_3 \times B(p,r)} \\+\alpha_4 \times P(p,r)+ \alpha_5 \times S(p,r), 
\end{array}\end{equation}
where $\alpha, \alpha_1, \alpha_2, \alpha_3, \alpha_4, \alpha_5$ indicate the weight coefficients of $N(p), G(p), V(p,r), B(p,r), P(p,r), S(p,r)$ respectively.

\subsection{Autonomous Evolution Image Fusion Network}
Based on the above research, we built an image fusion network with autonomous evolution ability. As shown in Figure \ref{general}, our network mainly consists of code base module, evaluation module and iterative optimization module. Among them, the code base module mainly includes image fusion methods of different image fusion tasks, which are used to generate fusion images including multi-exposure image fusion task, infrared and visible image fusion task, multi-focus image fusion task, medical image fusion task and remote sensing image fusion task. The evaluation module is mainly used to obtain the relative optimal solution of each iteration step for each type of image fusion task. The iterative optimization module mainly uses the relative optimal solution of each step obtained by the evaluation module to reverse optimize the main task of image fusion. In the training process, we need to use the code base module and evaluation module for each batch size image in the iteration process to obtain the relative optimal solution required by each iteration step. The weight training of neural network is inversely optimized by an optimal solution of bachsize. This is time-consuming in the first training process because of the need to solve a variety of image fusion methods. Therefore, in order to speed up the second and even the nth iteration training speed, we will reserve the initial solution of all image fusion methods and the relative optimal solution obtained by the evaluation function at the first time. Therefore, when it is necessary to add a new method to strengthen the network development, we only need to evaluate the optimal solution and the method and update the weight of the optimal solution.

In the main network of image fusion, our main network of image fusion will adopt the modular design, and adopt the pre trained image fusion weight to carry on the transfer learning directly. On the one hand, it can accelerate the convergence speed of image fusion, and only a small number of samples are needed to obtain robust results in the face of new tasks. On the other hand, we can replace different image fusion networks to enhance the network's ability in one aspect. It plays an important role in improving the self evolution ability of the whole network architecture. In this paper, we use the pre-training image fusion network as initial fusion weight \cite{fang2019crossmodale}. In the training phase, our hyper parameter settings are shown in Table \ref{talbe1}. 

\begin{table}[!h]
	
	\centering \footnotesize
	\renewcommand \arraystretch{1.1}
	\caption{
		Network hyperparameters
	}
	\label{table111}
	\begin{tabular}[b]{p{1cm}p{1cm}p{1cm}p{1cm}p{1cm}}
		\hline
		Type   &   Batchsize & Epoach & Learning rate & Image size\\ \hline
		Method &  5/10  &  255&0.00001&256/128\\ \hline
	\end{tabular}
	\label{talbe1}
\end{table}
Since our method can be updated and evolved in the image fusion module, we will not give the fixed image fusion network parameters here. If you want to obtain the pre-training fusion weight network of our paper, please refer to \cite{fang2019crossmodale}.

\section{Experiments}
\label{setup}
In this section, experimental setup are presented and comparative experiments result produced along with relevant explanations and analysis experiment are presented.

\subsection{Experimental Setup}
In this section, datasets, metrics and methods for experimental evaluation are first presented. Then, implementation details of evaluated methods are introduced.

\textit{1) Datasets:} \textit{a) Multi-exposure} \cite{JianruiCai2018LaDS}: This dataset contains 589 elaborately selected high-resolution multi-exposure sequences with 4,413 images.

\textit{b) Multi-focus} \cite{Nejati2015Multi}: For multi-exposure image fusion task, we will test it on MFIF data benchmark and \cite{Nejati2015Multi} dataset. MFIF dataset includes a test set of 105 image pairs, a code library of 30 MFIF algorithms, and 20 evaluation metrics. \cite{Nejati2015Multi} dataset contains 20 pairs of color multi-focus images of size $520\times520$ pixels and four series of multi-focus images with three sources.

\textit{c) Medical} \cite{Summers2003Harvard}: It includes 97 CT and MRI images and 24 T1-T2 weighted MRI images. The relevant images have registered the data.

\textit{d) Infrared and visible} \cite{zhang2020vifb,LiChenglong2019RotB, FLIR}: We will perform infrared and visible image fusion on FLIR dataset and VIFB dataset. FLIR dataset is obtained by RGB and thermal imaging camera installed on the vehicle, and contains 14, 452 thermal infrared images, including 10, 228 from short video and 4, 224 from 144 second video. Unfortunately, there is no registration. VIFB dataset includes 21 image pairs, 20 fusion algorithms and 13 evaluation indexes, which can be used for performance comparison. Fortunately, 20 algorithms corresponding to 21 images provide fused images. Unfortunately, no specific code has been released.

\textit{e) Combined vision system} \cite{Summers2003Harvard}: The dataset is specially used in the field of aviation visual navigation, including synthetic visual images and enhanced visual images. The dataset consists of 4000 pairs of original images. Through the fusion of synthetic and enhanced visual images, the combined visual images are obtained.

\begin{table}[!h]
	
	\centering \footnotesize
	\renewcommand \arraystretch{1.1}
	\caption{
		Experimental datasets and inherited properties
	}
	\label{table11}
	\begin{tabular}[b]{p{2cm}p{1cm}p{2cm}p{0.5cm}p{1cm}}
		\hline
		Type & Dataset   & Modality & Align & Matching pairs\\ \hline
		Multi-exposure \cite{JianruiCai2018LaDS}  &   \cite{JianruiCai2018LaDS} &  Multi-exposure  &  \checkmark&4, 413\\ 
		Multi-focus \cite{Nejati2015Multi, ZhangXingchen2020MIFA}  & Lytro and MFIF   &Multi-focus  &\checkmark& 125 \\ 
		Medical \cite{Summers2003Harvard}  & Brain    & CT, MRI  &\checkmark&97    \\ 
		Infrared and visible \cite{zhang2020vifb, FLIR, LiChenglong2019RotB}  & FLIR, VIFB and RGBT  & Infrared and visible  &  \checkmark& 14, 473  \\ 
		
		Combined vision system images (OURS) & CVS  & Enhanced and synthetic vision image  &  $\times$&4, 000  \\ 
		\hline
	\end{tabular}
\end{table}

The main properties of experimental datasets are summarized in Table. \ref{table11}. To evaluate the robustness of our framework, we performed experimental evaluations on different image fusion task datasets. 
\begin{table*}[!h]
	
	\centering \footnotesize
	\renewcommand \arraystretch{1.1}
	\caption{Metrics}
	\label{table122}
	\begin{tabular}[b]{p{0.05cm}p{1.3cm}p{7cm}p{6cm}}
		\hline
		No.& Method &Equations   & Description \\ 
		\hline
		1&EN \cite{1576816}&$EN=-\sum_{i=0}^{255} p_{i} \log _{2} p_{i},$&where $P_i$ is the probability of a gray level appearing in the image.\\ 
		\hline
		2&AG \cite{Cui2015Detail}&$AG=\frac{1}{M^{*} N} \sum_{i=1}^{M} \sum_{j=1}^{N} \sqrt{\frac{\Delta I_{x}^{2}(i, j)+\Delta I_{y}^{2}(i, j)}{2}},$& where $M \times N$ denotes the image height and width; $\Delta I_{x}(i, j)$ denotes image horizontal gradient; $\Delta I_{y}(i, j)$ denotes image vertical gradient.\\ 
		\hline
		3&SSIM \cite{1284395}& $\begin{array}{l}
		{SSIM}(I_i, R)=\frac{\left(2 u_{I_i} u_{R}+C_{1}\right)\left(2 \sigma_{I_i R}+C_{2}\right)}{\left(u_{I_i}^{2}+u_{R}^{2}+C_{1}\right)\left(\sigma_{I_i}^{2}+\sigma_{R}^{2}+C_{2}\right)},\end{array}$&where $\mu_{I_i}$ and $\mu_R$ indicate the mean value of origin image $I_i$ and fused image $R$; $\sigma_{I_iR}$ is the standard covariance correlation.\\ 
		\hline
		4&VIFF \cite{Han2013A}& $\mathrm{VIF}=\frac{\sum_{j \in  { subbands }} I\left({C} \stackrel{N, j}{;} {F}^{N, j} | s^{N, j}\right)}{\sum_{j \in \ { subbands }} I\left({C}^{N, j} ; {E}^{N, j} | s^{N, j}\right)},$& where ${C} \stackrel{N, j}{;}$ denotes N elements of the $C_j$ that describes the coefficients from subband j; $\sum_{j \in  { subbands }} I\left({C} \stackrel{N, j}{;} {F}^{N, j} | s^{N, j}\right)$ denotes reference image information.\\  
		
		\hline
		5&NIQE \cite{MittalA2013MaCB}& $\begin{array}{l}
		D\left(\nu_{1}, \nu_{2}, \Sigma_{1}, \Sigma_{2}\right) \\ \quad=\sqrt{\left(\left(\nu_{1}-\nu_{2}\right)^{T}\left(\frac{\Sigma_{1}+\Sigma_{2}}{2}\right)^{-1}\left(\nu_{1}-\nu_{2}\right)\right)}\end{array},$&where $\nu_{1}$ and $\Sigma_{1}$ are the mean vectors and covariance matrices of the natural MVG model and the distorted image’s MVG model.\\  
		
		\hline
		
		6&PSNR \cite{SijbersJ1996Qaio} &$\mathrm{PSNR}=10 \log _{10} \frac{\left(2^{n}-1\right)^{2}}{M S E},$& where $MSE$ is the mean square error of the current image X and the reference image y.\\ 
		
		\hline
		
		7&MI   \cite{Qu2002Information}&$\mathrm{MI(I_i,R)}=H(I_i)+H(R)-H(I_i, R),$&
		where $H(I_i)$ and $H(R)$ represent the information entropy of origin image and fused image; $H(I_i,R)$ denotes joint information entropy.\\   
		
		\hline
		8&Combined &$E_2 \ref{Eq2}$&	$E_2$ represents a variety of indicators to jointly represent image quality. For specific definition, refer to Eq. \ref{Eq2}.\\ 
		\hline
	\end{tabular}
\end{table*}

\textit{2) Metrics:} The distinctiveness of an image quality is usually quantitatively evaluated using entropy (EN) \cite{1576816}, average gradient (AG) \cite{Cui2015Detail}, structural similarity (SSIM) \cite{1284395}, visual information fidelity (VIF) \cite{Han2013A}, natural image quality evaluator  (NIQE) \cite{MittalA2013MaCB}, PSNR. \textit{(1) EN} \cite{1576816} represents information entropy. Information theory points out that the higher the information entropy is, the better the image quality is. \textit{(2) AG} \cite{Cui2015Detail} represents average gradient. It reflects the change rate of small detail contrast and represents the relative clarity of the image. Generally speaking, the higher the evaluation gradient, the higher the image level. \textit{(3) SSIM} \cite{1284395} denotes structureal similarity. The image quality is evaluated from three aspects: brightness, contrast and structure. The mean value is used as the estimation of brightness, the standard deviation as the estimation of contrast, and the covariance as the measurement of structural similarity. \textit{(4) VIF} \cite{Han2013A} represents visual Information Fidelity. The image quality is evaluated by simulating the significant physiological and psychological visual characteristics of the human visual system (HVS). \textit{(5) NIQE \cite{MittalA2013MaCB}} indicates natural image quality evaluator. \textit{(8) PSNR} denotes Peak Signal to Noise Ratio. The bigger the PSNR value, the better the image quality. \textit{(7) MI \cite{Qu2002Information}} represents mutual information. It indicates the correlation between two images. The more similar the images are, the greater the mutual information is. \textit{(8) Combined \ref{Eq2}} denotes combined evaluation criterion. The higher the combined value, the better the image quality.

\begin{table*}[!h]
	
	\centering \footnotesize
	\renewcommand \arraystretch{1.1}
	\caption{PARAMETER SETTINGS OF EVALUATED METHODS \cite{fang2019nonlinear}}
	\label{table12}
	\begin{tabular}[b]{p{0.05cm}p{1.8cm}p{4cm}p{0.6cm}p{2cm}p{1cm}p{1cm}p{1.7cm}p{1.5cm}}
		\hline \\
		No.& Method &Parameters &Year  & Category  &Time(s)&Evolution&Robustness and Generality &Expansibility \\ \hline 
		1&FZL \cite{Lahoud2019FastZERO}&$r_b = 35, \varepsilon_b = 0.01, r_d = 7, \varepsilon_d = 1e-6$&2019&Hybrid&0.56&$\times$&$\checkmark$&$\times$ \\ 
		2&CSR \cite{Liu2016ImageCSR}&$\lambda=0.01$ &2016&Multi-scale& 98.14&$\times$&$\times$&$\times$ \\ 
		3&DL \cite{Li_2018DL}& $\alpha_1=\alpha_2=0.5, k\subset[1,2]$&2018& Deep learning &18.62&$\times$&$\times$&$\times$\\ 
		4& DENSE \cite{Li2018DenseFuse}&$Epoach=4, Lr=0.0001$ &2019& Deep learing& 0.83&$\times$&$\times$&$\times$\\ 
		5&FusionGAN \cite{Ma2018Infrared}&$Epoach=10, Lr=0.0001$ &2019&GAN&0.10&$\times$&$\times$&$\times$\\ 
		6&IFCNN \cite{ZHANG202099}&$L_{r0}=0.01, power=0.9$&2020&Deep learning & 0.08&$\times$&$\checkmark$ &$\times$\\
		7&DTCWT \cite{Lewis2007Pixel}&$\times$&2007& Wavelets & 0.25&$\times$&$\times$&$\times$\\ 
		8&LATLRR \cite{Li2018InfraredLTLRR}&$\lambda =0.4, stride=1$ &2020&Multi-scale  &271.04&$\times$&$\times$&$\times$\\ 
		9&LP-SR \cite{Liu2015ALPSR}&$overlap=6, \epsilon =0.1, level=4$&2015&Hybrid& 0.04&$\times$&$\times$&$\times$\\ 
		10&DSIFT \cite{efae}&$Scale=48, blocksize=8, matching=1$&2015&Other& 3.98 &$\times$&$\times$&$\times$\\ 
		11&CNN\_IV \cite{Liu2017InfraredCNN} & $t=0.6$ &2017&Hybrid& 31.76 &$\times$&$\times$&$\times$\\ 
		12&CNN\_MF \cite{Liu2017MultiCNN} & $Momentum=0.9, decay=0.0005, threshold=0.5$ &2017&Hybrid& 31.76 &$\times$&$\times$&$\times$\\ 	
		13&CVT \cite{Nencini2007RemoteCVT} &$is_real=1, finest=1$&2007&Multi-scale & 1.09&$\times$&$\times$&$\times$\\ 
		14&CBF \cite{Shreyamsha2015ImageCBF} &$\sigma s=1.8, \sigma r=25, ksize=11$&2015&Multi-scale&22.97&$\times$&$\times$&$\times$\\ 
		15&JSR \cite{Zhang2013Dictionary}&$Unit=7, step=1, dic_size=256, k=16$	&2013&Sparse representation& 93.89&$\times$&$\times$&$\times$\\ 
		16&JSRSD \cite{Liu2017InfraredJSR-SD} &$Unit=7, step=1, dic_size=256, k=16$&2017&Saliency-based& 172.44 &$\times$&$\times$&$\times$\\ 
		17&GTF \cite{Ma2016InfraredGTF}&$Epsr=epsf=tol=1, loops=5$ &2016&Other & 6.27 &$\times$&$\times$&$\times$\\ 
		18&WLS \cite{Ma2017InfraredWLS}&$\sigma_s=2, \sigma_r=0.05, nLevel=4$&2017&Hybrid& 8.18&$\times$&$\times$&$\times$\\ 
		19&RP \cite{Toet1989ImageRP}&$\times$&1989&Pyramid&0.76 &$\times$&$\times$&$\times$\\ 
		20&MSVD \cite{Naidu2011Image}  &$\times$&2011&Multi-scale& 1.06&$\times$&$\times$&$\times$\\ 
		21&MGFF \cite{Durga2019Multi}&$R=9, \varepsilon=10^3, k=4$&2019&Multi-scale  & 1.08&$\times$&$\times$&$\times$\\
		22 &ZCA \cite{Li2018Infrared}& $K=2, i=4 and i=5$  &2019&Hybrid&2.57 &$\times$&$\times$&$\times$\\
		23&ADF \cite{Bavirisetti2016Fusion} &$w1=w2=0.5$&$2016$&Multi-scale &1.00&$\times$&$\times$&$\times$\\
		24&FPDE \cite{Bavirisetti2017Multi} &$At=0.9, n=20, k=4, \delta t=0.9$&$2017$&Subspace & 2.72&$\times$&$\times$&$\times$\\
		25&IFEVIP \cite{Zhang2017Infrared} &$Nd = 512, Md = 32, Gs = 9, MaxRatio = 0.001, StdRatio = 0.8 $&$2017$&Other & 0.17&$\times$&$\times$&$\times$\\	
		26&PGMI \cite{PGMI} &$Epoach=15, lr=1e-4, c_dim=1, stride=14, scale=3$&$2020$&Deep learning&0.30 &$\times$&$\times$&$\times$\\
		27&FusionDN \cite{xu2020aaai} &$Ps=64, lam=80000, num=40$&$2020$&Deep learning & 1.95&$\times$&$\checkmark$&$\times$\\
		28&SAF \cite{fang2019crossmodale} &$Epoach=256, bs=5, lr=0.000001$&$2020$&Deep learning &0.33 &$\times$&$\times$&$\times$\\
		29&AE-Net &Dynamic&$2020$&Deep learning &0.11&$\checkmark$&$\checkmark$&\checkmark\\
		\hline
	\end{tabular}
\end{table*}

\textit{3) Methods:} As shown in Table \ref{table12}, we will compare experiments with 29 mainstream algorithms such as fast-zero-learning (FEZ) \cite{Lahoud2019FastZERO}, fonvolutional sparse representation (CSR) \cite{Liu2016ImageCSR}, deep learning (DL) \cite{Li_2018DL}, generative adversarial network for image fusion (Fusion GAN) \cite{Ma2018Infrared}, dual-tree complex wavelet transform (DTCWT) \cite{Lewis2007Pixel}, latent low-rank representation
(LATLRR) \cite{Li2018InfraredLTLRR}, multi-scale transform and sparse representation (LP-SR) \cite{Liu2015ALPSR}, dense sift
(DSIFT) \cite{efae}, convolutional neural network (CNN) \cite{Liu2017InfraredJSR-SD}, curvelet transformation (CVT) \cite{Nencini2007RemoteCVT}, bilateral filter fusion method (CBF) \cite{Shreyamsha2015ImageCBF}, cross joint sparse representation (JSR) \cite{Zhang2013Dictionary}, joint sparse representation with saliency detection (JSRSD) \cite{Liu2017InfraredJSR-SD}, gradient transfer fusion (GTF) \cite{Ma2016InfraredGTF}, weighted least square optimization (WLS) \cite{Ma2017InfraredWLS}, a ratio of low pass pyramid (RP) \cite{Toet1989ImageRP}, multi-resolution singular value decomposition (MSVD) \cite{Naidu2011Image}, proportional maintenance of gradient and intensity (PGMI) \cite{PGMI}, densely connected network for image fusion (FusionDN) \cite{xu2020aaai},  Multi-modal Image Restoration and Fusion (CUNET) \cite{DengXin2020DCNN}, subjective attention image fusion (SAF) \cite{fang2019crossmodale} and AE-Net. 

\textit{4) Implementation details:} 
Before the experiment, we need to clarify the following questions.
\textit{1)} In all subsequent experiments, we converted all images into grayscale images for subsequent image fusion. In the experiment, we only need to train the model on a mixed dataset, not on a separate dataset. But in order to show the robustness and generality of our method, we only train on the mixed dataset, and test on multiple datasets separately. \textit{2)} In the experiment, in addition to the comparison experiment on VIFB dataset benchmark, the benchmark fusion results are used, and other experiments are tested by the code published in the paper. \textit{3)} In addition, we need to point out that the time efficiency of different algorithms is tested on the VIFB dataset. \textit{4)} Because there are too many image fusion methods and huge workload, in this paper, we will select the latest two image fusion methods in each image fusion task to build our algorithm library unit. In the future, we will continue to add more image fusion methods for testing to continuously improve the performance of our network architecture.  \textit{5)} Because the number of many datasets is very small, from dozens to tens of thousands, so in the follow-up experiments of this paper, we use the pre-training image fusion weight \cite{fang2019crossmodal} to retrain. Our experimental platform is desktop 3.0 GHZ i5-8500, RTX2070, 32G memory.

\begin{figure}[ht]
	
	
	\includegraphics[scale=1,width=0.5\textwidth]{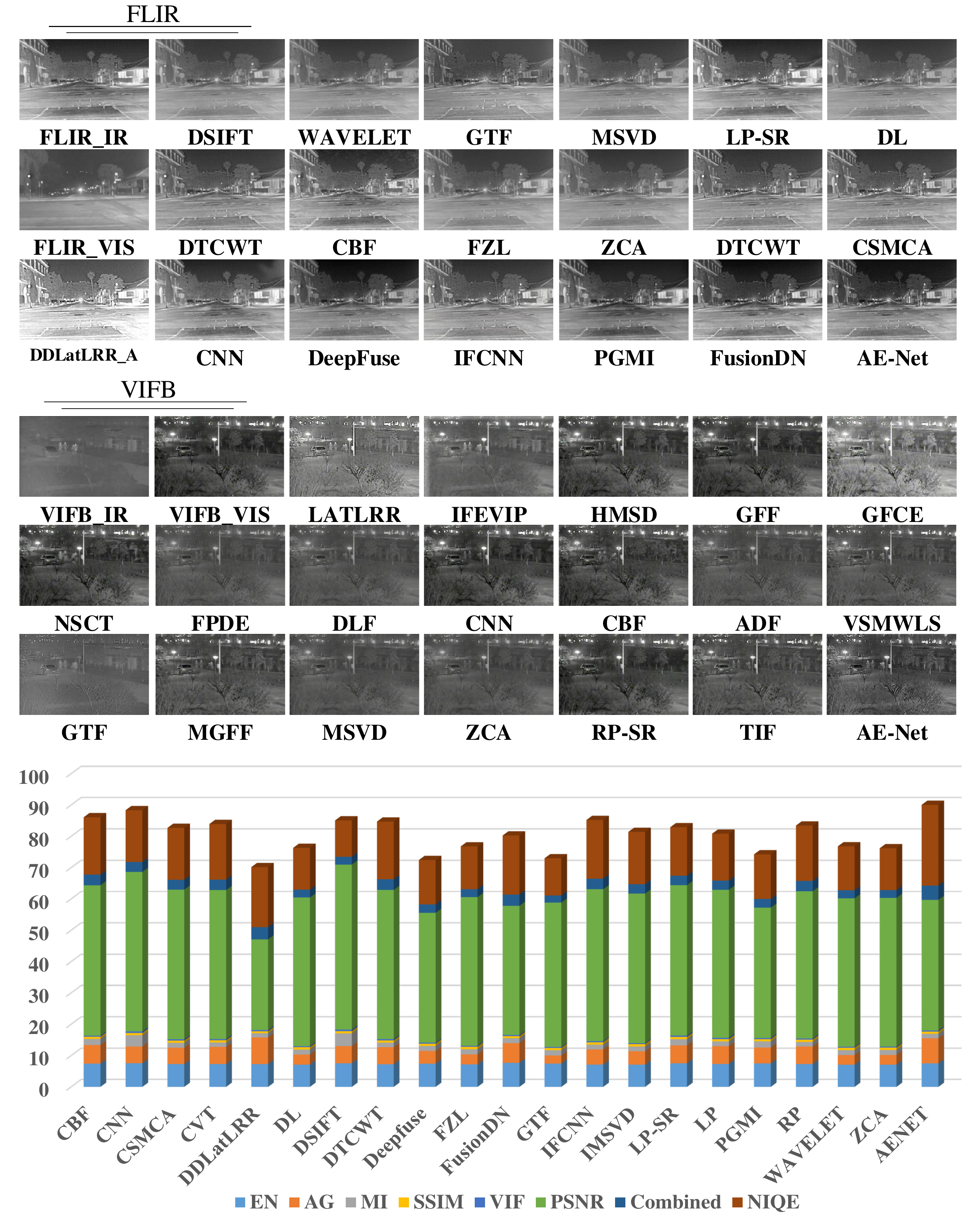}
	\centering
	
	\caption{
		Image fusion results of tested methods on infrared and visible images.}
	\label{f5}
\end{figure}
\subsection{Comparative experiments}
In this section, in order to verify the robustness and universality of our method, we will carry out comparative experiments and visual display on multi-exposure dataset, multi-focus dataset, medical dataset, infrared and visible dataset, combined vision system dataset. 

\subsubsection{Infrared and Visible Image Fusion}

Our image fusion method is tested on FLIR and VIFB datasets. Because there are too many image fusion methods, we only use some classic traditional ones methods and deep learning ones to test. The experimental results are shown in Figure \ref{f5}. From the Figure \ref{f5}, we can see that the latest image fusion methods, such as PGMI, FusionDN and IFCNN have better image fusion effect than existing methods. These three image fusion methods are compared with our method subjectively, we can find that although the gap is not big, whereas our image fusion effect in detail has better clarity. Even on the basis of VIFB dataset, our image fusion effect is the best. In order to further verify the effectiveness of our method, we also carried out quantitative analysis of objective indicators. We can see that for the full reference indicators SSIM, PSNR, and MI, our indicators are not absolutely dominant, whereas our method has absolute advantages in information entropy, gradient and comprehensive evaluation indicators. Because the full reference image is mainly used to evaluate the image data with ground truth labels, whereas for the cross-modal image data without ground truth labels, these indicators lack of effective representation ability, so we cannot evaluate the image quality simply by looking at the level of the full reference index. In view of this problem, \cite{fang2019crossmodal} also made a lot of research and exploration, and indicated the correctness of this point of view. For cross-modal image fusion data, human subjective evaluation is more representative than existing objective indicators.

\subsubsection{Multi-focus Image Fusion}

\begin{figure}[ht]
	
	
	\includegraphics[scale=1,width=0.47\textwidth]{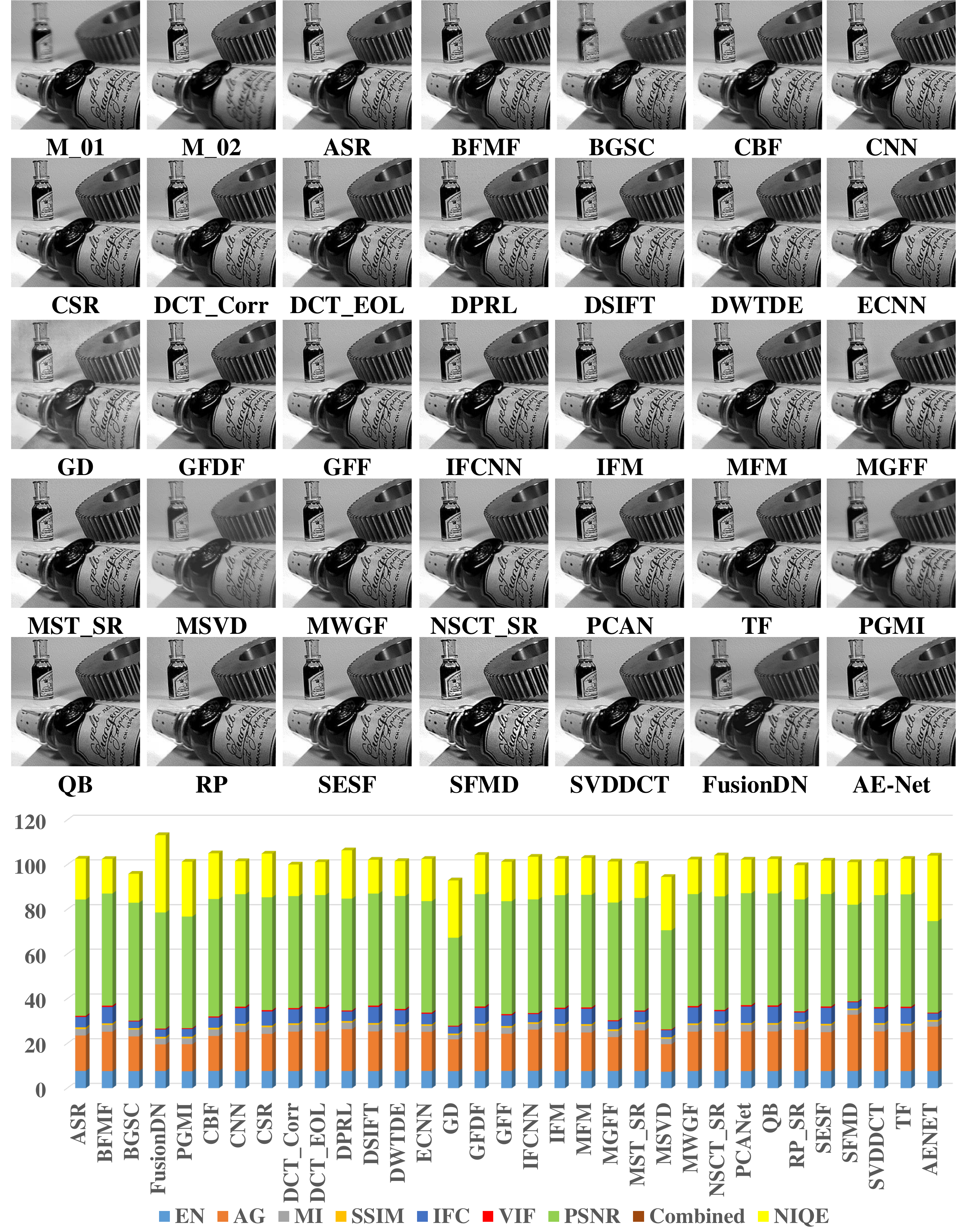}
	\centering
	
	\caption{
		Image fusion results of tested methods on multi-focus images.}
	\label{f6}
\end{figure}
Due to the problem of focal length and depth of scene of the camera, the images taken have different focus points. The full focus image can be formed by multi-focus image fusion. In this experiment, our training set and test set come from Lytro image fusion dataset and MFIF benchmark \cite{Nejati2015Multi, ZhangXingchen2020MIFA} respectively. Although from the perspective of subjective visual effects, CSR, ASR, BFMF, CNN, DCT\_EOL, DSIFT, IFCNN and AE-Net methods have very good clarity and high subjective quality. From the objective index point of view, the overall index of these image fusion methods is maintained at about 100, whereas FusionDN operator has higher objective index. On the contrary, the image quality of fusiondn is a little fuzzy. There are two main reasons for the analysis. \textit{There are two main reasons for the analysis. On the one hand}, we have not found the ground truth labels of the dataset, and existing image quality evaluation indicators cannot fully and effectively represent the image quality. \textit{On the other hand}, in the multi-focus fusion task, our AE-Net fusion effect is not the best, mainly because in the multi-focus module, our algorithm library only uses DSIFT, IFCNN operators in the training stage, which limits our image fusion effect to a certain extent. However, due to the autonomous evolution ability of our method, our image fusion performance will continue to improve with the increase of algorithm library.

\subsubsection{Multi-exposure Image Fusion}
\begin{figure}[ht]
	\includegraphics[scale=0.9,width=0.45\textwidth]{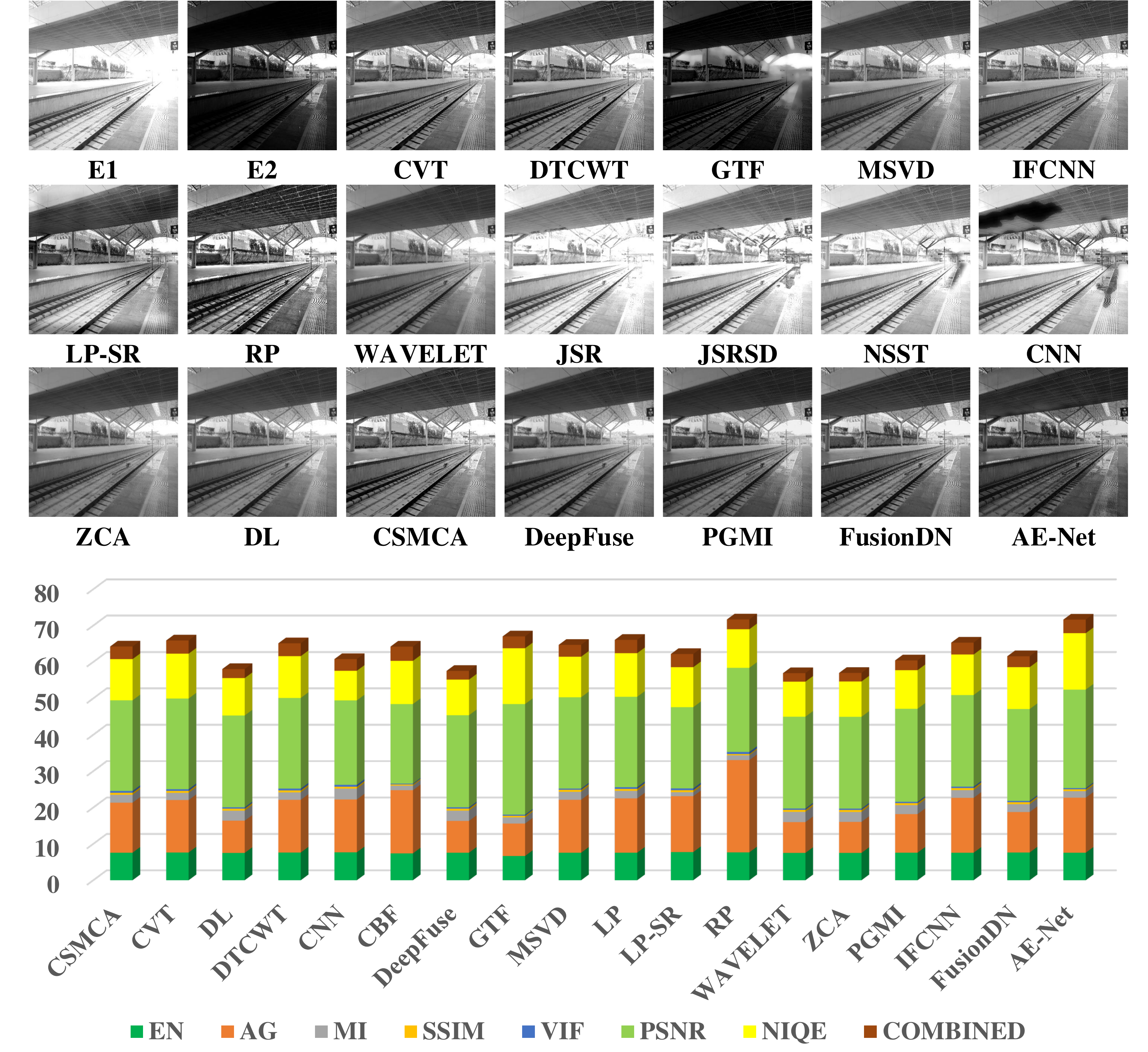}
	\centering
	
	\caption{
		Image fusion results of tested methods on multi-exposure images.}
	\label{f7}
\end{figure}

Due to the huge difference of target brightness, the difference leads to the problem of over exposure or under exposure in the captured image. Through the fusion of over exposure image and under exposure image, the image with high dynamic range can be formed. From Figure \ref{f7}, we can find that RP and our method has achieved outstanding results in both subjective evaluation and objective indicators. However, from the perspective of image subjective quality, RP introduces a lot of noise information to a certain extent, while still can maintain good clarity. However, our fusion effect keeps high definition and has less noise. We can also find CNN \cite{Liu2017MultiCNN} algorithm has obvious fusion oscillation problem. The main reason is that this method is mainly used for multi-focus image fusion task. Compared with the latest FusionDN, PGMI and IFCNN in $2020$, our image fusion results also have better advantages. On the one hand, these methds are part of our algorithm library. On the other hand, our method obtains the relative optimal solution in each step for reverse optimization.

\subsubsection{Combined Vision System Image Fusion}
\begin{figure}[ht]
	
	
	\includegraphics[scale=0.9,width=0.48\textwidth]{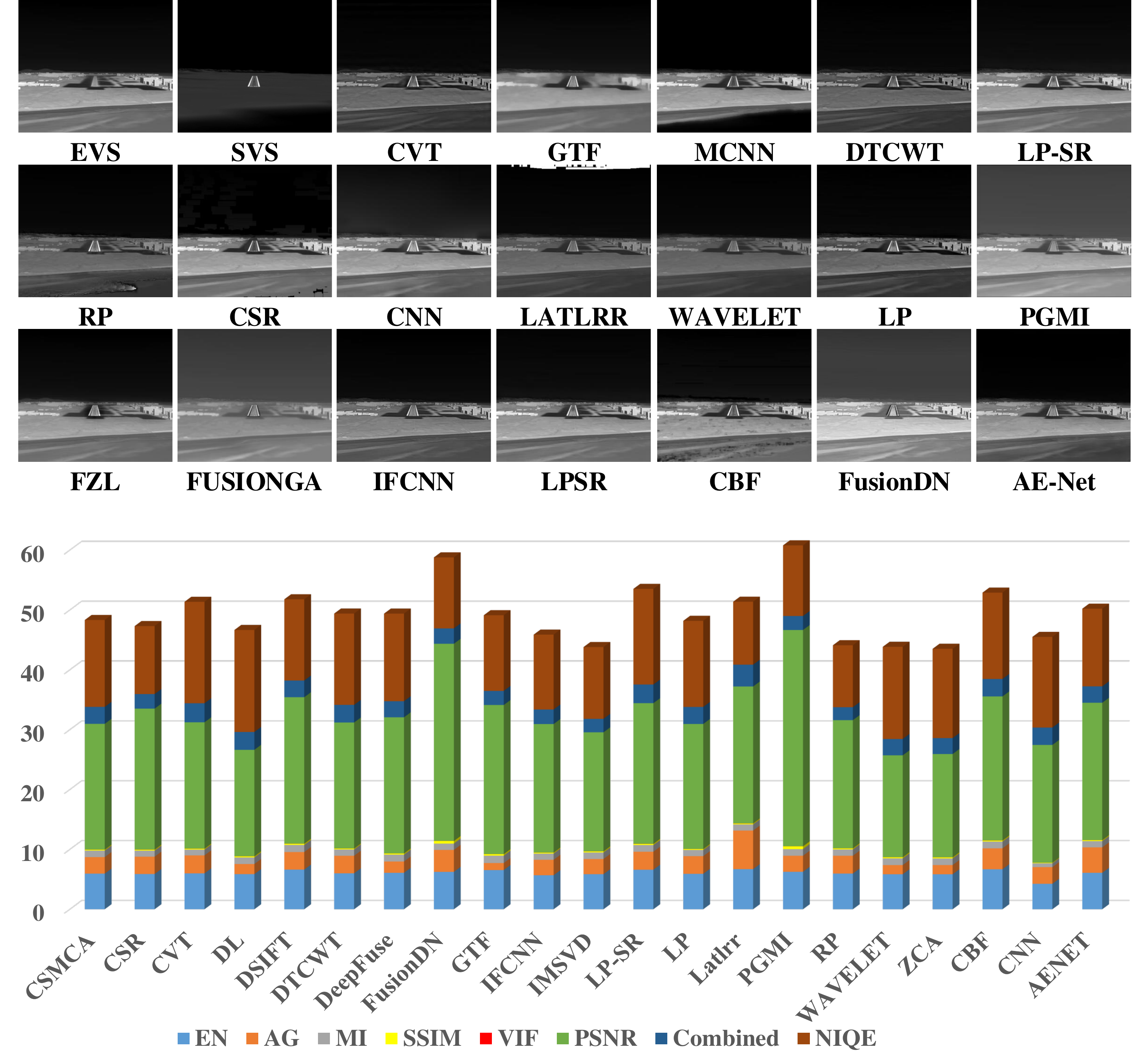}
	\centering
	
	\caption{
		Image fusion results of tested methods on CVS images.}
	\label{f8}
\end{figure}

The task of combined visual navigation is a new generation of aviation visual navigation technology in the future. It obtains SVS and enhanced data EVS data by collecting terrain data and atmospheric data on large aircraft, and obtains CVS image by fusing SVS and EVS data, and guides aircraft landing through CVS. At present, there are few related researches in this field, and there is no effective comparison method. Therefore, we use traditional visual fusion method and deep learning method for comparative analysis. From Figure \ref{f8}, we can find that DSIFT, PGMI, FusionDn, Latlrr and CBF have great advantages in objective image quality assessment. However, it is not difficult to find out that the main reason is that these image fusion operators have different degrees of fusion vibration. In particular, the highest index of fusiondn and pgmi operator, although the PSNR value is very high, whereas the image subjective quality is very low. Although LP-SR method has better performance than other operators in both subjective and objective image quality, we can clearly see that there is still a lot of brightness oscillation problem when we zoom in the image. In addition, there is a certain degree of ambiguity in the image clarity. From the gradient of objective indicators, we can also confirm that our method has higher performance than LP-SR operator The gradient of the value.

\subsubsection{Medical Image Fusion}
\begin{figure}[ht]
	
	
	\includegraphics[scale=0.9,width=0.48\textwidth]{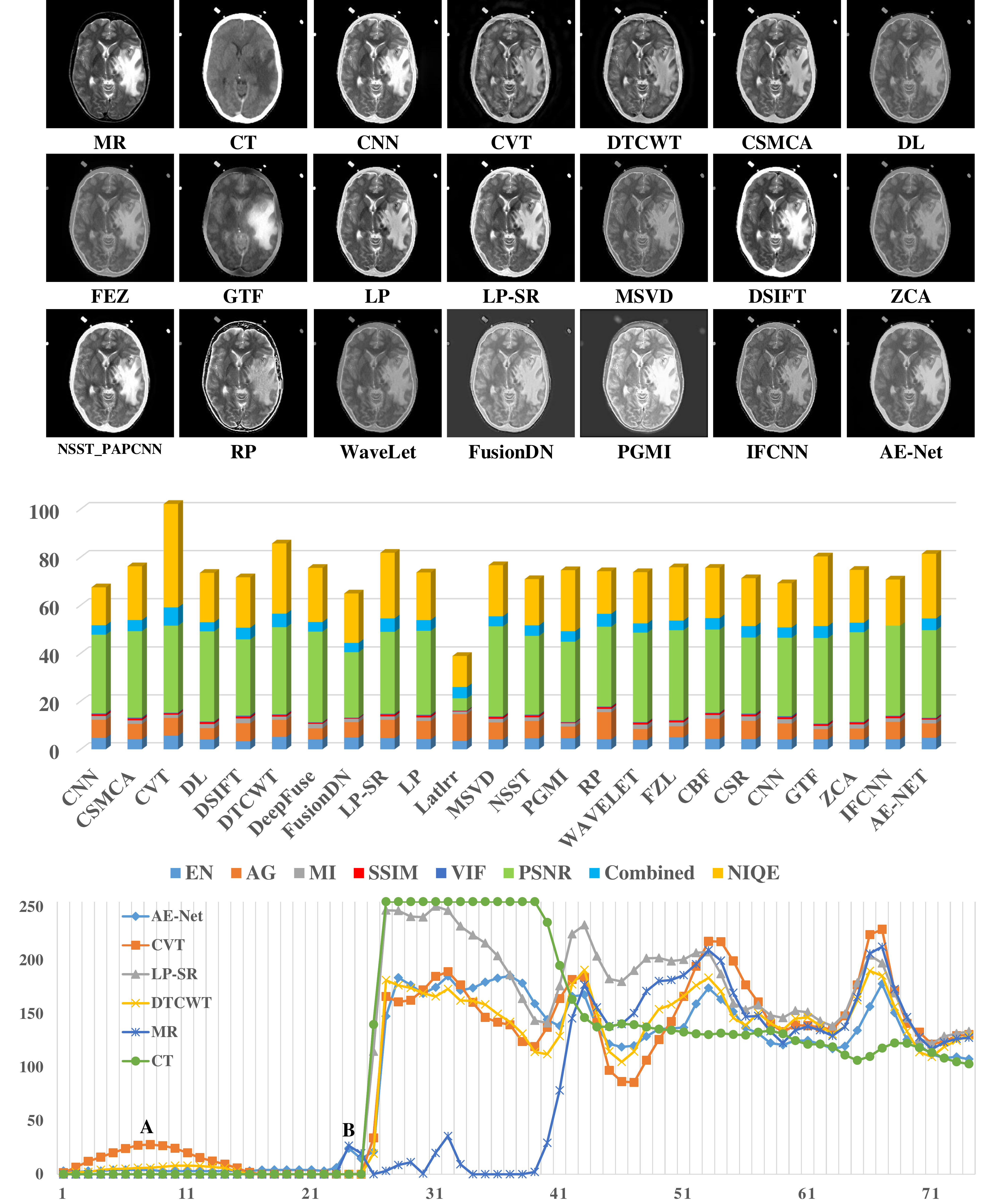}
	\centering	
	\caption{
		Image fusion results of tested methods on medical images.}
	\label{f9}
\end{figure}
The fusion of MR and CT images plays an important role in improving the diagnostic success rate of human brain diseases. In the task of medical image fusion, we mainly use CSMCA, IFCNN, PGMI. From Figure \ref{f9}, we can find that CVT, DTCWT, GTF, LP-SR methods have higher comprehensive evaluation index, whereas the original image will have obvious vibration, and the problem of missing information. We will mark the related problems in the image with infrared. Our algorithm fully retains the texture details of MR and CT images, and has better subjective visual effect. Compared with PGMI, CSMCA and IFCNN have better subjective and objective performance. Because the details of medical images are not obvious, we analyze the light intensity of four image fusion methods with higher objective index. The cross section coordinates of light intensity are \textit{$(0, 127) - (200, 127)$}. From the line graph, we can see that CVT method produces severe gray level oscillation effect at point A. DTCWT is also produced, but it is not obvious compared with CVT. In the detail of point B, CVT, DTCWT and LP-SR are missing obvious detail edge information. However, AE-Net effectively retains all the edge details, which is very important for medical image fusion task.

\subsection{Analysis Experiments}
At the same time, we also carry out analysis experiments for our method. 
\textbf{Firstly}, performance analysis experiment of autonomous evolution. \textbf{Secondly}, comparative analysis experiments using different loss function. \textbf{Finally}, comparative analysis of different autonomic evolution network architecture depth and time efficiency.

\subsubsection{Performance Analysis Experiment of Autonomous Evolution}

\begin{figure}[ht]
	
	
	\includegraphics[scale=0.9,width=0.49\textwidth]{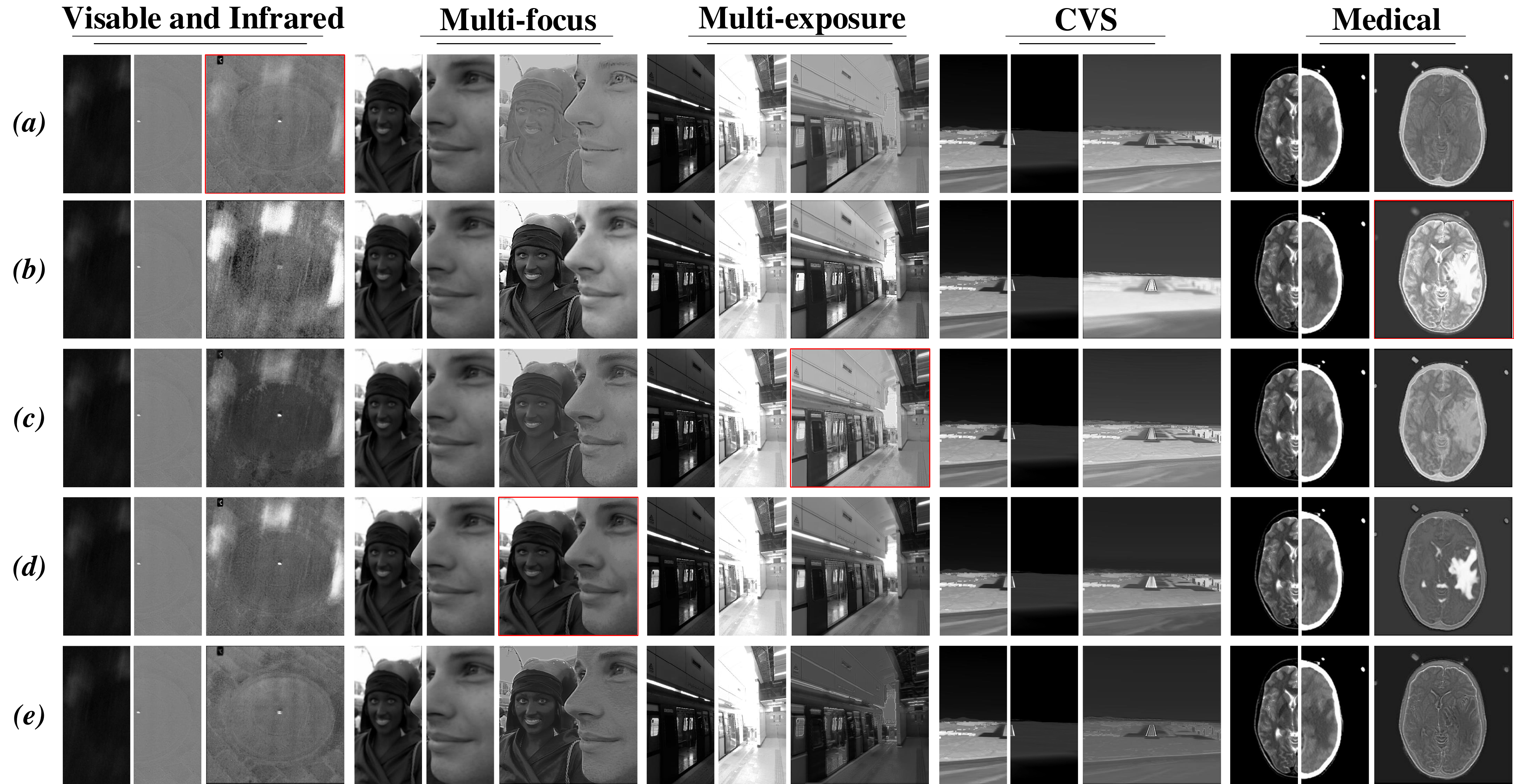}
	\centering
	\caption{PGMI algorithm for migration learning on different datasets. (a) Visible and infrared model. (b) Medical model. (c) Multi-exposure model. (d) Multi-focus model. (e) Pan shapening model.}
	\label{f10}
\end{figure}
Through extensive verification of the above algorithms in different image fusion tasks, we can find that existing image fusion methods have certain robustness and generality for different datasets. \textit{This kind of robustness and generality is based on direct transfer learning, that is to say, when training on one dataset, and then facing different datasets, it is necessary to add new datasets on this basis for fine tuning}. For example, CU-Net \cite{DengXin2020DCNN}, PGMI, DeepFuse et al. This kind of transfer learning makes the fusion effect on the target dataset very robust. Unfortunately the model is applied to the original image dataset fusion task again, the fusion effect will be greatly reduced. \textit{The problem of memory forgetting caused by transfer learning is the disadvantage of existing deep learning methods}.

\begin{figure}[ht]
	
	
	\includegraphics[scale=0.9,width=0.48\textwidth]{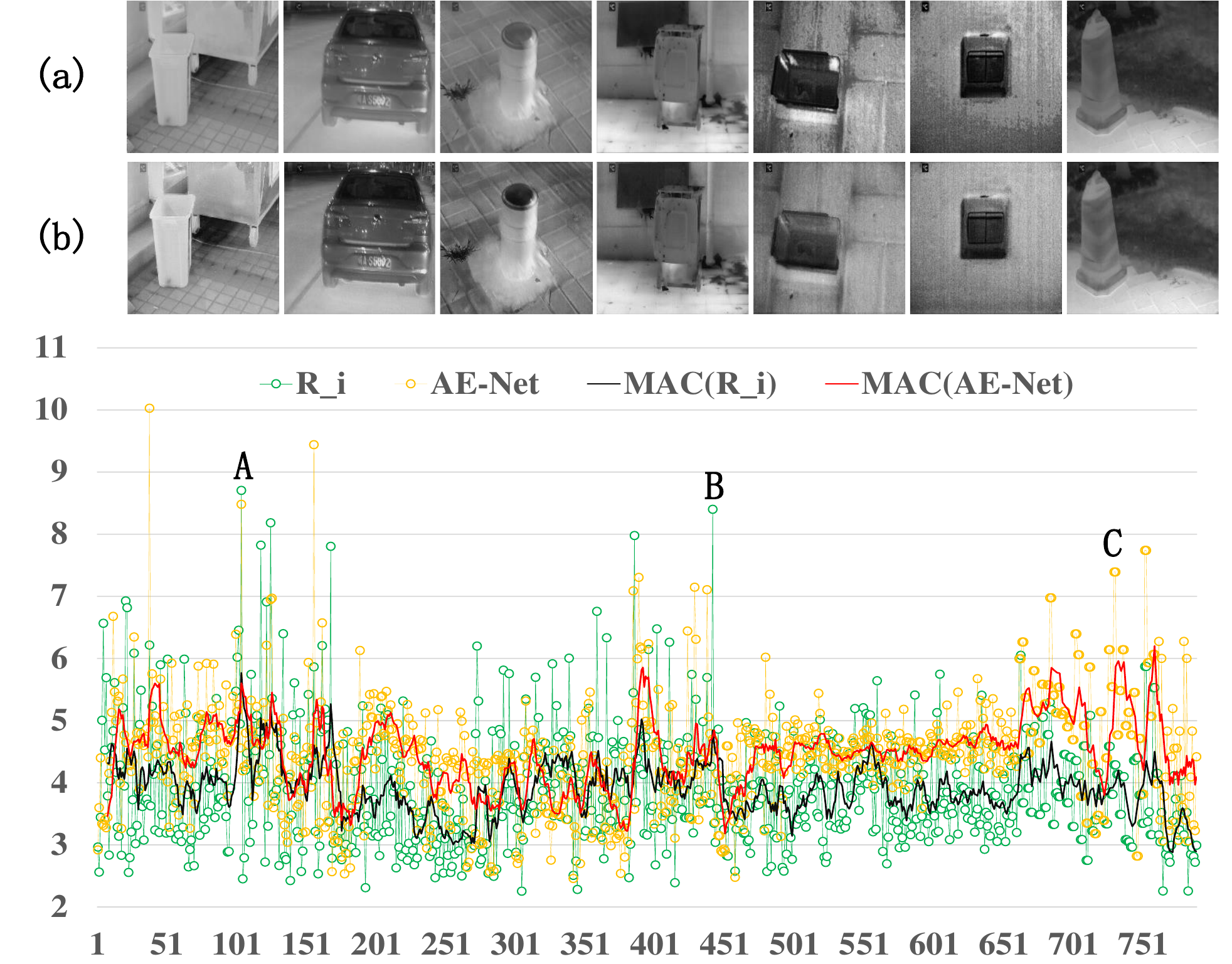}
	\centering
	
	\caption{Comparative experiment between relative optimal solution and AE-Net. (a) Relative optimal solution. (b) AE-Net image fusion result.}
	\label{f11}
\end{figure}

\textit{In order to verify the above problems, we designed a contrast experiment}. In the experiment, we will take $2020$ PGMI image fusion method as an example for comparative analysis. From the Figure \ref{f10}, we can see that the problem of memory forgetting caused by transfer learning is obvious. This also confirms the above viewpoint of this paper. \textit{In order to show the superiority of our method, in this experiment, we will compare the relative optimal solution with the fusion result of our image fusion method.} From Figure \ref{f11}, we can find two points. \textbf{Firstly}, although we use multiple metrics to evaluate the relative optimal solution, due to the lack of perfect evaluation metrics, the relative optimal solution maybe is wrong, which shows that the image contrast has changed greatly. \textbf{Secondly}, by learning the common features of multiple tasks, our method can alleviate the problem of false initial solution caused by image quality evaluation index to a certain extent. From the first line of images, we can see that due to the imperfection of evaluation metrics, the brightness contrast of the image changes dramatically, resulting in serious gray-scale distortion of the image. In the second row of images obtained by AE-Net, the concussion effect was effectively alleviated. The line graph in the third line represents the real-time comparison and trend comparison analysis between the relative optimal solution and AE-Net. The line graph in the third line represents the real-time comparison and trend comparison analysis between the relative optimal solution and AE-Net. The higher the index, the better the image quality. The horizontal axis represents the number of images. Point A indicates that the relative optimal solution is similar to the AE-Net, the change is small. Point B indicates that there is an error in the relative optimal solution obtained, and the objective index is very high. The corresponding original image has a sharp change in gray level. However, AE-Net index value is low, which avoids this kind of error. Point C indicates that the relative optimal solution obtained is wrong, which makes the image details incomplete. However, the index of AE-Net is higher, which indicates that AE-Net overcomes the error and retains more detailes and texture information.

\subsubsection{Comparative Analysis Experiments Using Different Loss Function}
The effectiveness of image quality assessment loss function has an important impact on image fusion. In this experiment, we make a detailed comparative analysis on the objective loss function of AE-Net. We mainly proceed from two aspects. \textit{We use the full supervision image quality evaluation loss function and the semi-supervision image quality evaluation loss function}. The supervised loss function refers to the optimal solution obtained by each step of evaluation as the ground truth labels. The semi-supervised method mainly combines the ground truth generated in each step and the input original image for joint evaluation and optimization. 
\begin{figure}[h]
	
	
	\includegraphics[scale=0.9,width=0.48\textwidth]{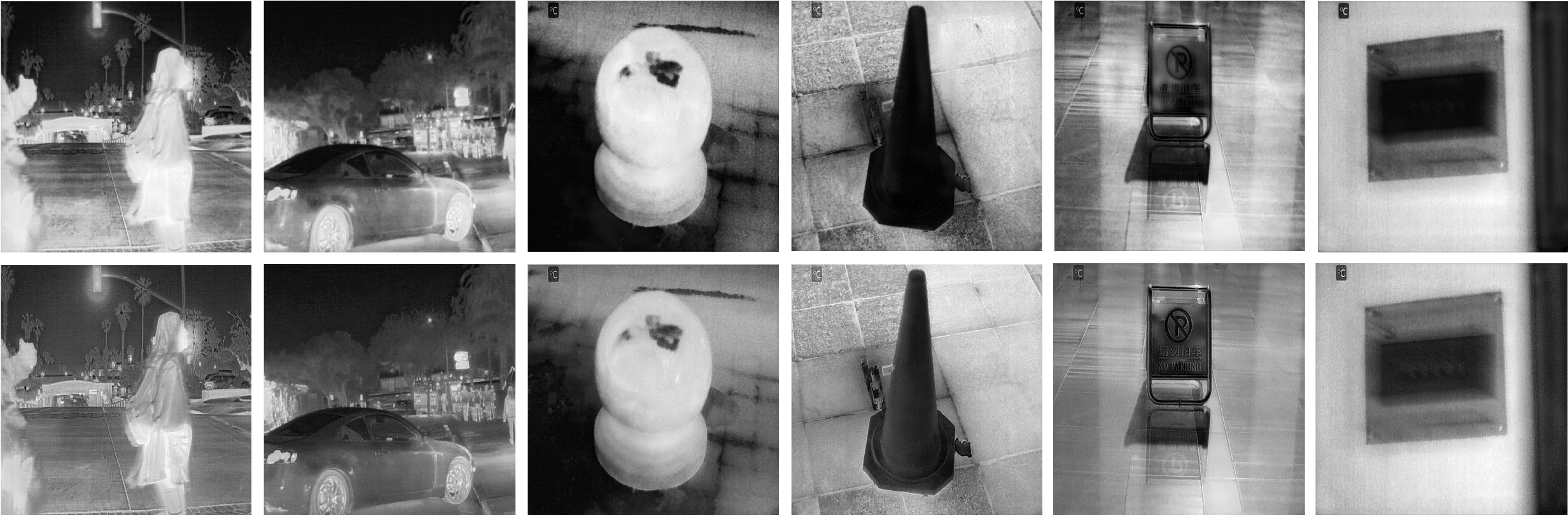}
	\centering
	
	\caption{Comparative analysis experiments using different loss function of AE-Net. Images are from RGBT and FLIR datasets.}
	\label{f12}
\end{figure}

From Figure \ref{f12}, we can find that the semi-supervised loss has better performance than the full supervision loss, and can alleviate the problem of poor image fusion quality caused by the error of the initial relative optimal solution to a certain extent. The main reason is that there are some errors in the relative optimal solution due to the imperfection of the evaluation function. If the supervision loss function is used completely, the network performance will be limited. However, even if we use complete end-to-end supervised learning, the image fusion quality has also achieved outstanding results.

\subsubsection{Comparative Analysis of Different AE-Net Architecture Depth and Time}
In order to analyze the performance of AE-Net method, we analyze the influence of different network depth on network performance and time. Due to the space limitation, we only change the network depth and calculate the time efficiency based on the SAF network architecture.  

\begin{figure}[h]
	
	\includegraphics[scale=0.9,width=0.48\textwidth]{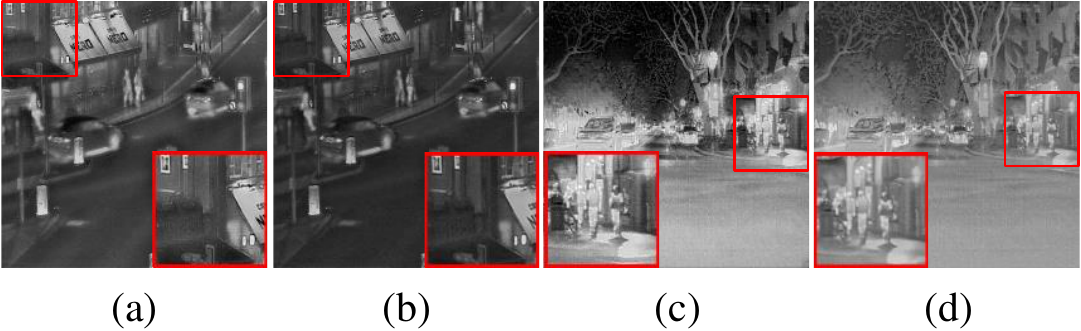}
	\centering
	\caption{AE-Net network size pruning operation. (a), (c) Before prune. (b), (d) After prune. Images are from TNO and FLIR datasets. }
	\label{f13}
\end{figure}
From the Table \ref{table12}, we can see that the time of image fusion is reduced to 3 times compared with SAF \cite{fang2019crossmodale}. From Figure \ref{f13}, there is a slight decrease in the gradient index, while the pruned network still retains the adaptability of AE-Net to complex scenes (high light, dark light). In this respect, existing image fusion methods cannot do. This is of great significance to the application and promotion of image fusion technology.


\section{Discussion}
Comprehensive experiments in section \ref{setup} verify that our image fusion method is better than existing image fusion methods in robustness and universality. This also prove the effectiveness of our simulation of human brain cognitive mechanism. We think there are several main reasons. 

\textit{1) Human Brain Cognitive Mechanism}. In the field of computer vision, the human brain has robustness and versatility for a variety of visual processing tasks, and its power consumption is absolutely superior to existing visual processing algorithms. The main reason is that the human brain has a perfect autonomous learning mechanism, which can continuously improve its ability, which is not possessed by existing visual processing algorithms. On the one hand, existing image fusion methods focus more on the improvement of the algorithm to improve the relevant image fusion indicators; on the other hand, existing image fusion algorithms lack of research and exploration of human brain cognitive mechanism. Therefore, we carry out the simulation of human brain cognitive mechanism in the field of image fusion for the first time. By introducing the human brain cognitive mechanism, the image fusion network has the ability of autonomous evolutionary learning of human brain. A large number of experimental results show that our simulation of human brain cognitive mechanism is effective, which has important guiding significance to improve the robustness and versatility of image fusion. This is the main reason why AE-Net chooses semi-supervised network architecture.

\textit{2) The Transformation From Unsupervised Learning to Semi-supervised Learning}. In the task of image fusion, it is very difficult to construct the objective function for unsupervised learning method which lacks label data. On the one hand, the nonlinear difference of image imaging attributes leads to the loss of similarity of local features and even deep convolution features. On the other hand, there is a lack of effective label data and perfect image quality evaluation index. Although many methods have been proposed for cross-modal image fusion task, such as image fusion method based on confrontation generation network or multi-task collaborative optimization method. However, the robustness and versatility of these image fusion methods are limited. To solve this problem, our proposed general image fusion network can transform unsupervised learning into supervised learning, which can improve the robustness and versatility of image fusion to a certain extent.


Our method effectively unifies the image fusion task with human brain cognitive mechanism, and effectively improves the continuous learning ability of image fusion method. Compared with existing image fusion methods, our image fusion method has obvious advantages in generality and robustness. \textit{\textbf{However}, the human brain is a very complex system. When processing visual tasks, human beings should not only make use of the continuous learning ability of human brain, but also make use of the brain's ability to understand image content}. This problem is not only the deficiency of our method, but also the lack of theoretical research on existing image fusion methods. \textit{\textbf{In addition}, the evolution of our method is dependent on the performance of different task domain algorithms. If the algorithm module provides some very bad image fusion algorithms, then the evolution will evolution in the bad direction, otherwise it will evolution in the right direction.}

\section{Conclusion}
Inspired by human brain cognitive mechanism, we proposed a robust and general image fusion method. \textit{The main differences between our image fusion method and existing image fusion methods are as follows}. \textbf{Firstly}, our image fusion method analyzes the cognitive mechanism of human brain for the first time, and establishes a physical model by simulating the human brain.
\textbf{Secondly}, by simulating the cognitive mechanism of human brain, our image fusion method has the function of autonomous evolution of human brain, and has the ability of continuous learning. \textbf{Finally}, our method is more robust and general than existing methods. 

Our image fusion method can be applied to various image fusion tasks. To verify the effectiveness of our image fusion network, we conduct exhaustive experiments on five different image fusion tasks. We also make a comprehensive ablation study to explore the consistency between the design of network architecture and the cognitive mechanism of human brain. In addition, we also explored the influence of different evaluation functions on image quality. The research we have done shows that our method outperforms other state-of-the-art methods in robustness and generality.

\begin{flushleft}	
\textbf{Acknowledgment}\\
\end{flushleft}
We are very grateful to Prof. Roundtree and Dr. Xiaoming Wang for their support of the language of the paper. This work was supported by the National Natural Science Foundation of China under Grants nos. 61871326, and the Shaanxi Natural Science Basic Research Program under Grant no. 2018JM6116.

\section*{References}

\bibliography{mybibfile}

\end{document}